\documentclass[envcountsame,oribibl]{llncs}

\usepackage[utf8]{inputenc} 
\usepackage[T1]{fontenc}    
\usepackage{hyperref}       
\usepackage{url}            
\usepackage{booktabs}       
\usepackage{amsfonts}       
\usepackage{nicefrac}       
\usepackage{microtype}      
\usepackage{color}

\usepackage{graphicx}
\usepackage{float}
\usepackage{subcaption}
\makeatletter
\newcommand{\figcaption}{\def\@captype{figure}\caption}
\newcommand{\tabcaption}{\def\@captype{table}\caption}
\makeatother
\usepackage{silence}
\usepackage{multirow}
\usepackage{amsmath}
\usepackage{bm}
\usepackage[square,sort,comma,numbers]{natbib}
\makeatletter
\renewcommand*{\@fnsymbol}[1]{\ensuremath{\ifcase#1\or \dagger\or \ddagger\or \ddagger\or
    \mathsection\or \mathparagraph\or \|\or **\or \dagger\dagger
    \or \ddagger\ddagger \else\@ctrerr\fi}}
\makeatother

\usepackage[nottoc,notlof,notlot]{tocbibind}

\begin{document}

\title{\large Deep De-Aliasing for Fast Compressive Sensing MRI}
\author{Simiao Yu\thanks{\scriptsize Co-first authors}\inst{1}, Hao Dong$^\dagger$\inst{1}, Guang Yang\inst{2}, Greg Slabaugh\inst{3}, Pier Luigi Dragotti\inst{4}, Xujiong Ye\inst{5}, Fangde Liu\inst{1}, Simon Arridge\inst{6}, Jennifer Keegan\inst{2}, David Firmin\inst{2}, \and Yike Guo\inst{1}\thanks{\scriptsize Corresponding author. Emails: \{simiao.yu13,hao.dong11,g.yang\}@imperial.ac.uk \protect\\ \{p.dragotti,fangde.liu,j.keegan,d.firmin,y.guo\}@imperial.ac.uk \protect\\ 
gregory.slabaugh.1@city.ac.uk, xye@lincoln.ac.uk, s.arridge@cs.ucl.ac.uk }}
\institute{Data Science Institute, Imperial College London, SW7 2AZ, London, UK
\and National Heart \& Lung Institute, Imperial College London, SW3 6NP, London, UK
\and Dept. of Computer Science, City University London, EC1V 0HB, London, UK
\and EEE Department, Imperial College London, SW3 6NP, London, UK
\and School of Computer Science, University of Lincoln, LN6 7TS, Lincoln, UK
\and CMIC, University College London, WC1E 6BT, London, UK
}
\maketitle
\setcounter{footnote}{0}

\begin{abstract}
\noindent Fast Magnetic Resonance Imaging (MRI) is highly in demand for many clinical applications in order to reduce the scanning cost and improve the patient experience. This can also potentially increase the image quality by reducing the motion artefacts and contrast washout. However, once an image field of view and the desired resolution are chosen, the minimum scanning time is normally determined by the requirement of acquiring sufficient raw data to meet the Nyquist–Shannon sampling criteria. Compressive Sensing (CS) theory has been perfectly matched to the MRI scanning sequence design with much less required raw data for the image reconstruction. Inspired by recent advances in deep learning for solving various inverse problems, we propose a conditional Generative Adversarial Networks-based deep learning framework for de-aliasing and reconstructing MRI images from highly undersampled data with great promise to accelerate the data acquisition process. By coupling an innovative content loss with the adversarial loss our de-aliasing results are more realistic. Furthermore, we propose a refinement learning procedure for training the generator network, which can stabilise the training with fast convergence and less parameter tuning. We demonstrate that the proposed framework outperforms state-of-the-art CS-MRI methods, in terms of reconstruction error and perceptual image quality. In addition, our method can reconstruct each image in 0.22ms--0.37ms, which is promising for real-time applications.
\end{abstract}

\section{Introduction}
\label{Introduction}

Magnetic Resonance Imaging (MRI) is an invaluable technique for clinical medical imaging in that it provides non-invasive, reproducible and quantitative measurements of tissue structural, anatomical and functional information. Despite its unique flexibility for imaging different tissue types and organs of the human body (because the sensitivity of the image to tissue characteristics can be extensively tuned), prolonged acquisition times limit its usage due to expensive cost and considerations of patient comfort and compliance \cite{Hollingsworth2015}. 

MRI is associated with an inherently slow acquisition speed that is due to data samples not being collected directly in the image space but rather in \textit{k}-space, which contains spatial-frequency information. Here \textit{k}-space and the image space are inversely related: resolution in one domain determines extent in the other. The raw data samples are acquired sequentially in \textit{k}-space and the speed at which \textit{k}-space can be traversed is limited by physiological and hardware constraints \cite{Lustig2008}. Once the desired field-of-view and spatial resolution of the MRI images are prescribed, the \textit{k}-space raw data we need to acquire is conventionally determined by the Nyquist–Shannon sampling criteria \cite{Nyquist1928}. 

One possible fast MRI approach is to undersample \textit{k}-space, to which can provide an acceleration rate proportional to the undersampling ratio. However, this undersampling in \textit{k}-space would violate the Nyquist–Shannon sampling criteria, and thus generates aliasing artefacts once the images have been reconstructed. Therefore, the main challenge for fast MRI is to find an algorithm that can reconstruct an uncorrupted or de-aliased image from the undersampled \textit{k}-space.

The mathematical framework of Compressive Sensing (CS) has been intensively investigated from about a decade ago \cite{Donoho2006}, and was almost immediately considered for fast MRI applications due to the inherent suitability of the MRI data. Firstly, in general, the medical imagery acquired by MRI is naturally compressible. CS utilises the implicit sparsity of MRI images to reconstruct accelerated acquisitions \cite{Fair2015}. Here the term `sparsity' describes a matrix of image pixels or raw data points, which is predominately zero valued or namely compressible. Such sparseness may exist either in the image domain or more commonly via a suitable mathematical representation in a transform domain of the images due to redundancy in a single image or over a series of related images. Using this property, CS allows accurate reconstruction from undersampled raw data, with the proviso that the sampling pattern is `random' to create incoherent undersampling artefacts that a proper `nonlinear reconstruction' can be applied to suppress noise-like artefacts without degrading image quality of the reconstruction \cite{Lustig2008}. Secondly, the MRI scanners acquire raw data samples in spatial-frequency encoded \textit{k}-space. This allows aforementioned random undersampling to be implemented on the scanner, and the fast MRI can be achieved by acquiring less data in the first place, although this violates the requirements of the Nyquist–Shannon sampling criteria.

Sparse regularisation, which is a key component for successful fast CS based MRI (CS-MRI), can be explored in a specific transform domain or generally in a dictionary-based subspace \cite{Lustig2007}. Classic fast CS-MRI used predefined and fixed sparsifying transforms, e.g., total variation (TV) \cite{Block2007,JunfengYang2010,Knoll2011}, discrete cosine transforms \cite{Hong2011,Lingala2013,Wang2014} and discrete wavelet transforms \cite{Qu2012,Zhu2013,Lai2016}. In addition, this has been extended to more flexible sparse representation learnt directly from data using dictionary learning \cite{Ravishankar2011,Caballero2014,Zhan2016}.  

Despite the promise of using fast CS-MRI, most routine clinical MRI scannings are still based on standard Cartesian sequences or accelerated only using parallel imaging. The main challenges are: (1) it can be challenging to satisfy the incoherence criteria required by CS \cite{Hollingsworth2015}; (2) the sparsifying transforms used in current CS methods might be too simple to capture complex image details associated with subtle differences of the biological tissues, e.g., TV based sparsifying transform can introduce staircase artefacts in the reconstructed image \cite{Yang2016}; (3) nonlinear optimisation solvers usually involve iterative computing and updating that may result in relatively long reconstruction time \cite{Hollingsworth2015}; (4) inappropriate hyper-parameters predicted in current CS methods can cause over-regularisation that will yield overly smooth and unnatural looking reconstructions or images with residual undersampling artefacts \cite{Hollingsworth2015} and (5) the acceleration rate is still limited (2$\times$ to 6$\times$ acceleration).

Recently, deep learning has been widely and successfully applied in many computer vision problems. Essentially, CS-MRI reconstruction is solving a generalised inverse problem that is analogous to image super-resolution (SR), de-noising and inpainting that have been well investigated in computer vision. Deep neural network architectures, especially convolutional neural networks (CNNs), are becoming the state-of-the-art technique for tackling such inverse problems, e.g., image SR \cite{Dong2016}, de-noising and inpainting \cite{Xie2012,Agostinelli2013}. Comprehensive reviews on classic CS methods and clinical applications can be found elsewhere \cite{Hollingsworth2015,Jaspan2015}, and here we briefly review the most relevant publications using deep learning models for the CS-MRI. It is of note that despite the popularity of deep learning in computer vision applications, there has only been preliminary research on deep learning based CS-MRI. From our literature search, we found only two formal publications on this topic \cite{Wang2016,Yang2016}, and the other three preprints on the arXiv \cite{Schlemper2017a,Hammernik2017,Lee2017}. In general these methods leveraged deep learning (e.g., CNNs) to derive an optimal mapping between the undersampled \textit{k}-space (aliased reconstruction) and the desired uncorrupted image (de-aliased reconstruction). This has been done in either a sequential manner with classic CS-MRI or in an integrated manner that considered the CNNs based training as an additional regularisation term \cite{Wang2016}. The experimental results have shown some promise; however, the improvement was not significantly different from what classic CS-MRI can achieve although the reconstruction speed has been dramatically improved \cite{Yang2016}. Moreover, only up to 6$\times$ acceleration could be achieved by these methods \cite{Wang2016,Yang2016,Schlemper2017a,Hammernik2017,Lee2017}.    

In this study, we proposed a novel conditional Generative Adversarial Networks (GAN) based deep learning architecture for fast CS-MRI. Our main contributions are: (1) we used the U-Net architecture \cite{Ronneberger2015} with skip connections to achieve better reconstruction details; (2) for more realistic reconstruction, we proposed to combine the adversarial loss with a novel content loss considering both pixel-wise mean square error (MSE) and perceptual loss defined by pretrained VGG networks and (3) we proposed a refinement learning for training the generator that can stabilise the training with fast convergence and less parameter tuning. Compared to other state-of-the-art CS-MRI methods, we can achieve up to 10$\times$ acceleration with superior results and faster processing time using the GPU.

\section{Method}
\label{Method}

\subsection{General CS-MRI}

MRI reconstruction naturally deals with complex numbers. Let $\mathrm{x}\in\mathbb{C}^N$ represent a complex-valued MRI image, which consists of $\sqrt{N}\times\sqrt{N}$ pixels formatted as a column vector. The aim is to reconstruct $\mathrm{x}$ from the undersampled \textit{k}-space measurements $\mathrm{y}\in\mathbb{C}^M$ ($M<<N$), such that $\mathrm{y} = \mathrm{F}_u \mathrm{x}$, in which $\mathrm{F}_u\in\mathbb{C}^{M \times N}$ is the undersampled Fourier encoding matrix. In order to solve this underdetermined and ill-posed system, one must exploit a-priori knowledge of $\mathrm{x}$ that can be formulated as an unconstrained optimisation problem, that is
\begin{equation}
\min_{\mathrm{x}}~\frac{1}{2} || \mathrm{F}_u \mathrm{x} - \mathrm{y} ||^2_2 + \lambda \mathcal{R}(\mathrm{x}),
\label{Eq:unconstrainedOpt}
\end{equation}
where the least squares part represents the data fidelity term. $\mathcal{R}$ expresses regularisation terms on $\mathrm{x}$ and $\lambda$ is a regularisation parameter. The regularisation terms $\mathcal{R}$ typically involve $l_q$-norms ($0\leq q \leq 1$) in the sparsifying domain of $\mathrm{x}$ \cite{Lustig2008}. 

Previous deep learning based fast CS-MRI studies \cite{Wang2016,Schlemper2017a} integrated CNNs into CS-MRI, that is
\begin{equation}
\min_{\mathrm{x}}~\frac{1}{2} || \mathrm{F}_u \mathrm{x} - \mathrm{y} ||^2_2 + \lambda \mathcal{R}(\mathrm{x}) + \zeta || \mathrm{x} - f_{\mathrm{cnn}}(\mathrm{x}_u | \hat{\theta}) ||^2_2,
\label{Eq:CNNs}
\end{equation}
in which $f_{\mathrm{cnn}}$ is the forward propagation of the CNNs parametrised by $\theta$, and $\zeta$ is another regularisation parameter. The image generated by the CNNs (i.e., $f_{\mathrm{cnn}}(\mathrm{x}_u | \hat{\theta})$) was used as a reference image and as an additional regularisation term, in which $\hat{\theta}$ represents the optimised parameters of the trained CNNs. In addition, $\mathrm{x}_u = \mathrm{F}_u^H \mathrm{y}$ is the reconstruction from the zero-filled undersampled \textit{k}-space measurements, where $H$ represents the Hermitian transpose operation.

\subsection{General GAN}
Generative Adversarial Networks (GAN)~\cite{Goodfellow2014} consist of a generator network $G$ and a discriminator network $D$. The goal of the generator $G$ is to map latent variable $\bm{z}$ to the distribution of the given true data in order to fool a discriminator $D$, while the discriminator aims to distinguish the true data $\bm{x}$ from the synthesised fake data $G_{\theta_G}(\bm{z})$. Mathematically, the training process can be represented by a minimax function with network parameters $\theta_G$ and $\theta_D$ as following
\begin{equation}
\min_{\theta_G}\max_{\theta_D}~\mathbb{E}_{\bm{x}\sim p_{\mathrm{data}}(\bm{x})}[\log D_{\theta_D}(\bm{x})]+\mathbb{E}_{\bm{z}\sim p_{\bm{z}}(\bm{z})}[\log(1-D_{\theta_D}(G_{\theta_G}(\bm{z})))],
\label{Eq:Gan}
\end{equation}
where latent variable $\bm{z}$ is sampled from a fixed latent distribution $p_{\bm{z}}(\bm{z})$ and real samples $\bm{x}$ come from a real data distribution $p_{\mathrm{data}}(\bm{x})$. Extra prior information can constrain the generator $G$ to learn to create samples conditioned on such information, which is known as conditional GAN \cite{Mirza2014}.

\subsection{Proposed Method}

\begin{figure}[!ht]
	\begin{center}
	 \includegraphics[width=1\textwidth]{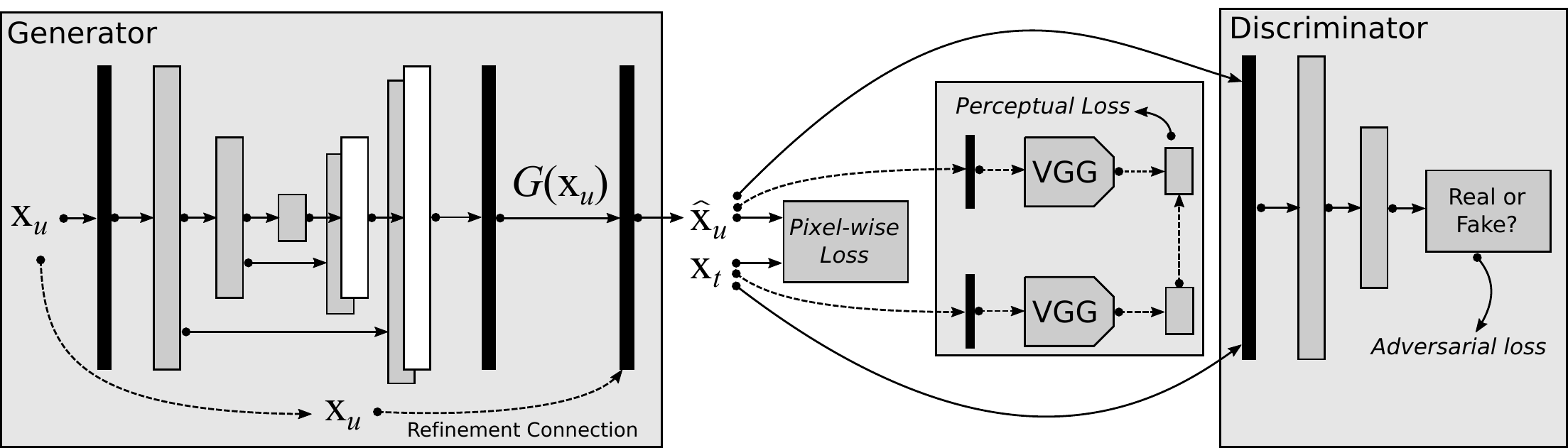}
	\end{center}
	\caption{Schema for our proposed GAN-based de-aliasing for fast CS-MRI.}
	\label{fig:Flowchart}
\end{figure}

Figure \ref{fig:Flowchart}~shows the overall framework of our GAN-based de-aliasing architecture. We used GAN conditioned on images. We fed the zero-filling reconstruction $\mathrm{x}_u$ (contains aliasing artefacts) to the generator and would yield the corresponding de-aliased reconstruction $\mathrm{\hat{x}}_u$ in order to stop the discriminator from recognising it from fully-sampled ground truth reconstruction $\mathrm{x}_t$. In other words, we have paired $\mathrm{x}_t$ and $\mathrm{x}_u$ as the input training data, we output $\mathrm{\hat{x}}_u$.

\textbf{Adversarial Loss}~~First, instead of using CNNs, we incorporated adversarial loss into our de-aliasing process for the fast CS-MRI that can be expressed as
\begin{equation}
\min_{\theta_G}\max_{\theta_D}~\mathbb{E}_{\mathrm{x}_t\sim p_{\mathrm{train}}(\mathrm{x}_t)}[\log D_{\theta_D} (\mathrm{x}_t)] + \mathbb{E}_{\mathrm{x}_u\sim p_{G}(\mathrm{x}_u)}   [\log (1-D_{\theta_D}(G_{\theta_G}(\mathrm{x}_u)))]. 
\label{Eq:AdversarialLoss}
\end{equation}

\textbf{Adversarial Learning}~~For the generator $G$, we used the U-Net architecture \cite{Ronneberger2015}, which applied skip connections between mirrored layers in the encoder and decoder paths. These skip connections can pass different levels of features to the decoder and gain better reconstruction details. For the output activation function of the generator, we used the hyperbolic tangent function. This adversarial learning process can be considered as using an adaptive loss function to iteratively shift the distribution of the de-aliased reconstruction towards the ground truth distribution. 

\textbf{Content Loss}~~In order to make the generated images more realistic, in addition to the adversarial loss, we also designed a content loss for training the generator, which is formed by coupling a pixel-wise MSE loss and a perceptual VGG loss, that is  
\begin{align}
    \min_{\theta_G}~\alpha\frac{1}{2} ||\mathrm{x}_t-\mathrm{\hat{x}}_u||^2_2 + \beta\frac{1}{2} || f_{\mathrm{vgg}}(\mathrm{x}_t) &- f_{\mathrm{vgg}}(\mathrm{\hat{x}}_u) ||^2_2 + \log (1-D_{\theta_D}(\mathrm{\hat{x}}_u))\label{Eq:MSEAndVGG}   \\ 
    \mathrm{s.t.}~~\mathrm{\hat{x}}_u &= G_{\theta_G}(\mathrm{x}_u), \notag
\end{align}
in which the first term describes the MSE loss that is a common choice of the optimisation cost function for the deep learning based fast CS-MRI \cite{Yang2016,Hammernik2017}. In our study, we used the normalised MSE (NMSE). However, the solution solely based on the optimisation of the MSE loss, which is defined on pixel-wise image difference, could result in perceptually overly smooth reconstructions that often lack high-frequency image details. We therefore defined an additional VGG loss (second term of Eq. \ref{Eq:MSEAndVGG}) to take the perceptual similarity into account \cite{Ledig2016}. In particular, we used the conv4 output of the VGG as the encoded embedding of the de-aliased output and the ground truth, and computed the MSE between them. By optimising this combined loss, the aim is to train a generator network successfully that can yield realistic de-aliased reconstruction that can fool the discriminator network. Once the generator network has been trained, we can apply it to any new inputs (i.e., initial aliased zero-filled reconstructions), and it will result in the de-aliased reconstruction.  

\textbf{Refinement Learning}~~Another main innovation of our method is that we added the undersampled reconstruction $\mathrm{x}_u$ to the generator output to model the final de-aliased reconstruction, i.e., instead of using $\mathrm{\hat{x}}_u = G_{\theta_G}(\mathrm{x}_u)$ we proposed to use $\mathrm{\hat{x}}_u =  G_{\theta_G}(\mathrm{x}_u)+\mathrm{x}_u$. In so doing, we transferred the generator from a conditional generative function to a refinement function, i.e., only generate the missing information. This can dramatically reduce the complexity of the learning, make the model more stable with faster convergence. In order to ensure that the de-aliased reconstruction $\mathrm{\hat{x}}_u$ is in a proper intensity scale as the ground truth, we applied a simple ramp function to rescale the image.        

Our generator can learn to perform de-aliasing of the zero-filled reconstruction, and create solutions that highly resemble the fully sampled ground truth. We named our method (using GAN architecture, pixel-wise MSE loss, VGG loss, and refinement learning) as \underline{P}ixel-\underline{P}erceptual-\underline{G}AN-\underline{R}efinement (PPGR). For comparison purpose, we also tested the method without refinement learning that is named as \underline{P}ixel-\underline{P}erceptual-\underline{G}AN (PPG), and the method with pixel-wise MSE only and GAN architecture is denoted as \underline{P}ixel-\underline{G}AN (PG).

\section{Experimental Settings and Results}
\label{Results}

\subsection{Experimental Settings}

\hspace{15pt} \textbf{Datasets}~~First, we trained and validated our GAN-based deep de-aliasing model using a subset of the IXI dataset \footnote{\scriptsize http://brain-development.org/ixi-dataset}. We randomly selected 1605 T\textsubscript{1}-weighted MRI images acquired for healthy volunteers. For this subset of the IXI dataset, we demonstrated the robustness of our model using 5-fold cross-validation. Second, we also tested our model using a MICCAI 2013 grand challenge dataset \footnote{\scriptsize  http://masiweb.vuse.vanderbilt.edu/workshop2013/index.php/Segmentation\_Challenge\_Details}. We randomly included 100 T\textsubscript{1}-weighted MRI data for training and 50 MRI data for testing as described by \cite{Yang2016} for a comparison study. We simulated both 1D and 2D Gaussian distribution based undersampling masks for our experiments. For each mask, 10\%, 20\%, 30\%, 40\% and 50\% remaining raw \textit{k}-space data were simulated representing 10$\times$, 5$\times$, 3.3$\times$, 2.5$\times$ and 2$\times$ accelerations.

\textbf{Evaluation Methods}~~For both datasets, we reported the NMSE, the Peak Signal-to-Noise Ratio (PSNR in dB) and the Structural Similarity Index (SSIM) \cite{Ledig2016}. The reconstructed fully sampled \textit{k}-space data was used as ground truth (GT) for validation. In addition to quantitative metrics, we also evaluated our method using qualitative visualisation of the reconstructed MRI images and the error with respect to the GT (e.g., using absolute difference image amplified 10$\times$).      

\textbf{Networks and Training Settings}~~Our GAN architecture was inspired by \cite{Radford2015, Isola2017} that consists of multiple convolutional and deconvolutional layers with batch normalisation \cite{Ioffe2015} and leaky ReLU layers. The VGG network \cite{Simonyan2015} we used was pretrained on ImageNet \cite{Russakovsky2015}. Detailed architecture settings can be found in the Supplementary Material. For both datasets, we trained separate networks for different sampling ratios with the following mutual hyperparameters: $\alpha=15$, $\beta=0.0025$, initial learning rate of 0.0001, batch size of 25. We adopted Adam optimisation \cite{KingmaAdam2014} with momentum of 0.5. For the IXI dataset, each model was trained by fixed 60 epochs and the learning rate was halved every 30 epochs. For the MICCAI dataset, each model was learnt by employing early stopping and the learning rate was halved every 5 epochs.

\textbf{Data Augmentation}~~The purpose of data augmentation is to improve the network performance by intentionally producing more training data from the original one. Conventional data augmentation (e.g., image flipping, rotation, shift, brightness adjustment and zoom) can result in displacement fields change but can not create training samples with diverse shapes. The shape of the organs imaged by MRI could be diverse but the variation is limited; therefore, in addition to conventional data augmentation, we also applied elastic distortion \cite{Simard2003} that can generate more training data with arbitrary but reasonable shape variations.

\textbf{Implementation}~~The implementation of our GAN-based de-aliasing model has been done using a high-level Python wrapper (TensorLayer \footnote{\scriptsize  http://tensorlayer.readthedocs.io}) of the TensorFlow \footnote{\scriptsize https://www.tensorflow.org} library. We will publish our open-source implementation on the Github.

\subsection{Results}

\begin{figure}[h]
	\begin{center}
	\includegraphics[scale=0.48, trim={0cm 0cm 0cm 0cm},clip]{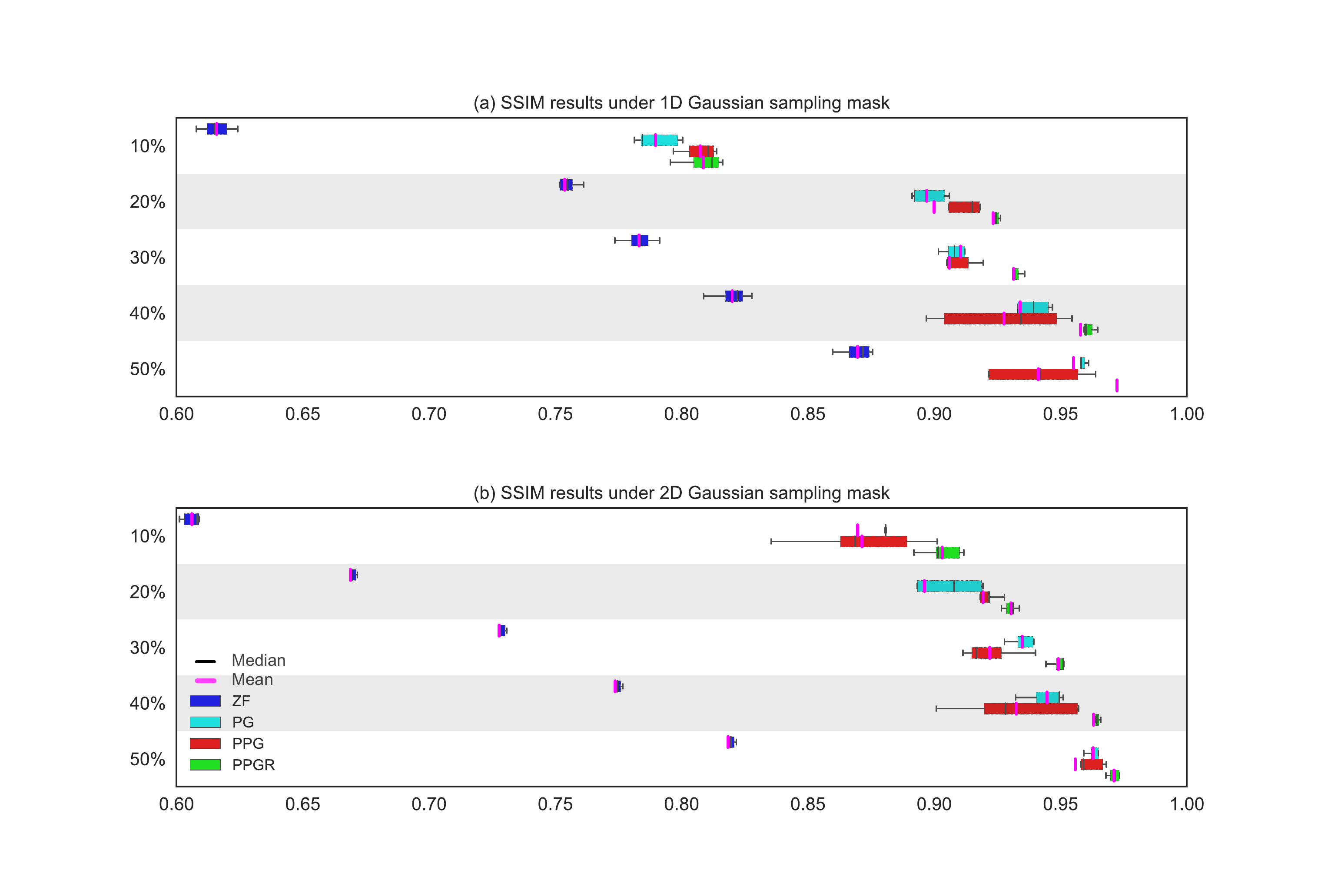}
	\end{center}
	\caption{\small\bf\it \small\bf\it Box plots of the SSIM for the reconstruction of the IXI dataset (a) using 1D Gaussian random sampling mask (for only the phase encoding direction) and (b) using 2D Gaussian random sampling mask.}
\label{fig:Quant_SSIM}
\end{figure}

Figure \ref{fig:Quant_SSIM}~shows comparison results of the SSIM between our proposed methods (PG, PPG and PPGR) and the baseline zero-filling (ZF) reconstruction using the IXI dataset (higher SSIM indicates better results). In general, all versions of our GAN-based methods outperformed the baseline reconstruction significantly. PPGR with refinement learning obtained the best SSIM with less variation than PPG and PG methods. We also calculated the NMSE and PSNR (refer to the Supplementary Material), both of which gained significant improvement compared to the baseline reconstruction regardless the random sampling distribution and undersampling ratio.

Table \ref{tab:MICCAI_quant}~tabulates the NMSE and PSNR results for our MICCAI dataset. Compared to the ZF baseline, our methods also performed much better. When only 10\% of \textit{k}-space remained using a 2D undersampling mask, we could still obtain $>38$dB PSNR. Compared to \cite{Yang2016} study that used a similar experimental setting and data, we achieved better reconstruction in terms of NMSE and PSNR improvements. Also, we demonstrated that our method could work when the \textit{k}-space is highly undersampled (e.g., when only 10\% of raw data remains we can still obtain SSIM $>0.97$). Figure \ref{fig:Quant_Convergence} (a) shows the line profile comparison using different reconstruction methods. Compared to the line profile of the GT image, ZF results were clearly over-smooth. Although better details can be observed in PG and PPG results compared to ZF, the PPGR results were the best that were closer to the GT. Figure \ref{fig:Quant_Convergence} (b) and (c) demonstrate that after adding the refinement connection, our PPGR method had a much faster convergence and a more stable improvement over the PSNR than the PPG. 

As the MICCAI datasets we used and the experimental settings (2D undersampling masks) were similar to \cite{Yang2016}, we also compared with their quantitative results. Compared to the conventional CS-MRI methods using predefined and fixed sparsifying transforms (e.g., TV \cite{Lustig2007}, RecPF \cite{JunfengYang2010} and PBDW \cite{Qu2012}), our PPGR method achieved lower NMSE and higher PSNR (e.g., $\mathrm{NMSE}\geq0.062$ using conventional CS-MRI v.s. $\mathrm{NMSE}=0.059$ using PPGR, and $\mathrm{PSNR}<39$dB v.s. $\mathrm{PSNR}=43.56$dB when the undersampling ratio is 30\%). Compared to state-of-the-art CS-MRI methods using non-local sparsity operator, dictionary learning and deep learning (e.g., PANO \cite{Qu2014}, FDLCP \cite{Zhan2016}, BM3D-MRI \cite{Eksioglu2016} and ADMM-Net \cite{Yang2016}), we obtained comparable NMSE and improved PSNR, and more importantly our GPU implementation could process each reconstruction in only 0.22ms--0.37ms. 
 
\begin{figure}[h]
	\begin{center}
	\includegraphics[scale=0.23, trim={0cm 0cm 0cm 0cm},clip]{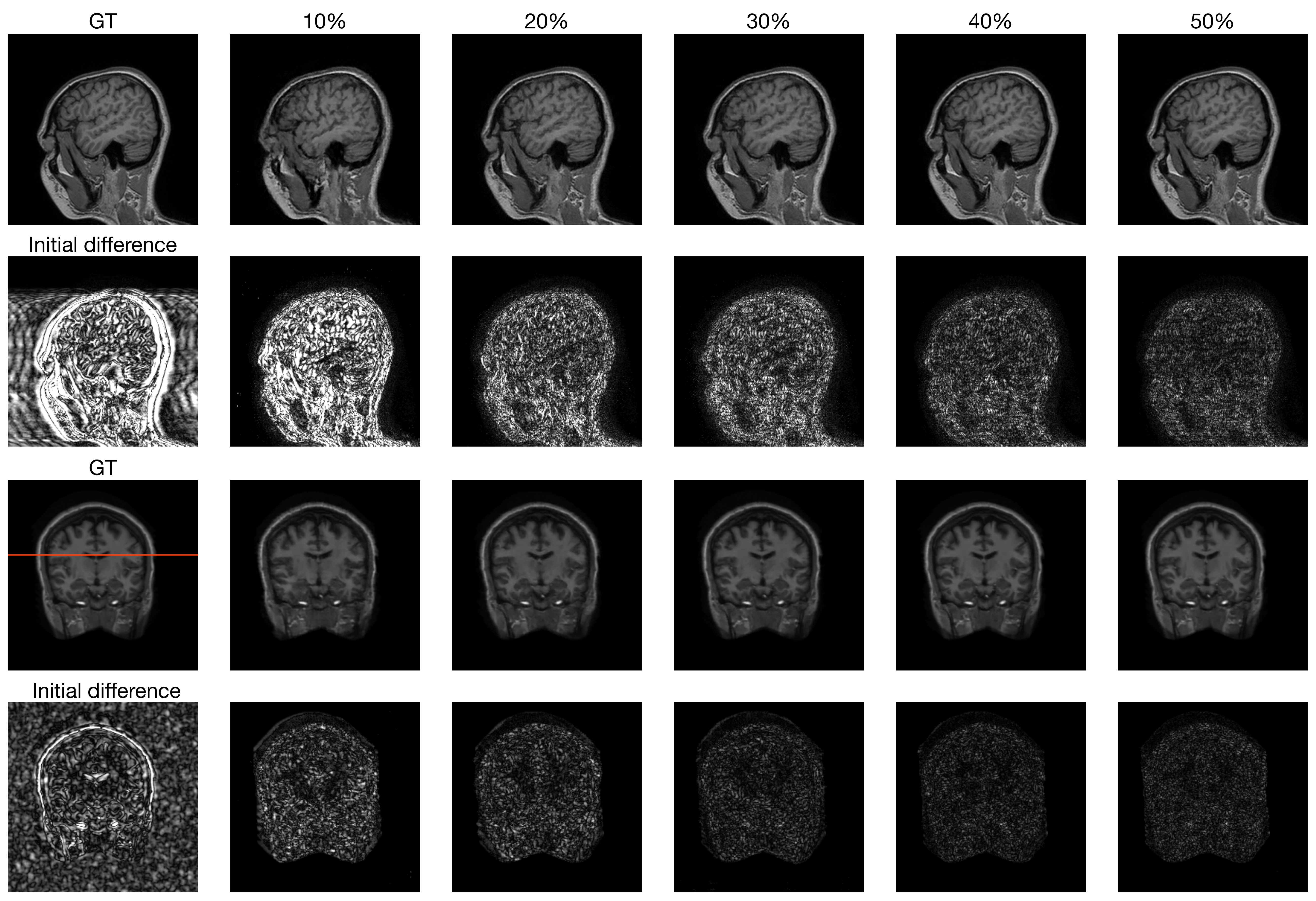}
	\end{center}
	\caption{\small\bf\it Qualitative visualisation of our PPGR de-aliasing for fast CS-MRI.}
\label{fig:Quali_CS-MRI}
\end{figure}

Figure \ref{fig:Quali_CS-MRI}~shows examples of reconstructed images using IXI dataset (upper two rows) under 1D Gaussian random sampling and MICCAI datasets (bottom two rows) under 2D Gaussian random sampling with various undersampling ratios, respectively. For each example case, we showed the GT and the reconstructions with various undersampling ratios (Figure \ref{fig:Quali_CS-MRI}~1st and 3rd rows), and the difference images between the GT and each reconstruction (Figure \ref{fig:Quali_CS-MRI}~2nd and 4th rows). We obtained compelling de-aliasing results compared to the initial difference images that contained significant aliasing artefacts. We can hardly observe any qualitative differences between the reconstructed images and the GT when the undersampling ratio $\geq$30\%. However, when the undersampling ratio $\leq$20\%, we might start noticing the loss of structural information (e.g., organ edges); however, most of the aliasing artefacts have still been suppressed effectively. Also, in general results obtained from 2D undersampling masks were better than the ones recovered using 1D masks.

\section{Discussion}
\label{Discussion}

Classic fast CS-MRI treats the image reconstruction task as a nonlinear optimisation problem without considering prior information of the expected appearance of the anatomy or the possible structure of the undersampling artefacts. This is significantly different from how human radiologists read images. Radiologists have been trained to read MRI images and scrutinise for certain reproducible anatomical and contextual patterns \cite{Hammernik2017}. By reading thousands of MRI images over the course of their career, they can obtain remarkable skills to understand images with known artefacts presented \cite{Hollingsworth2015}. Therefore, imitating this human learning experience using GAN based deep learning model can change the conventional online nonlinear optimisation task into an offline training procedure. In other words, compared to classic CS methods that solved the inverse problem for each new input dataset, our GAN based deep learning model can learn the complex nonlinear mapping between the undersampled \textit{k}-space (aliased reconstruction) and the desired uncorrupted image (de-aliased reconstruction) offline. Once such optimal mapping is learnt, it can be applied to any new input dataset. Although the training procedure can take a long time to finish (depending on the size of the training dataset and the desired quality of the reconstructed image), the reconstruction for the new input dataset is really fast (average 0.22ms per image on a Nvidia TitanX Pascal GPU) that is suitable for real time applications.

\begin{figure}[h]
    \centering
    \begin{subfigure}[b]{0.5\textwidth}
        \includegraphics[height=2.3in]{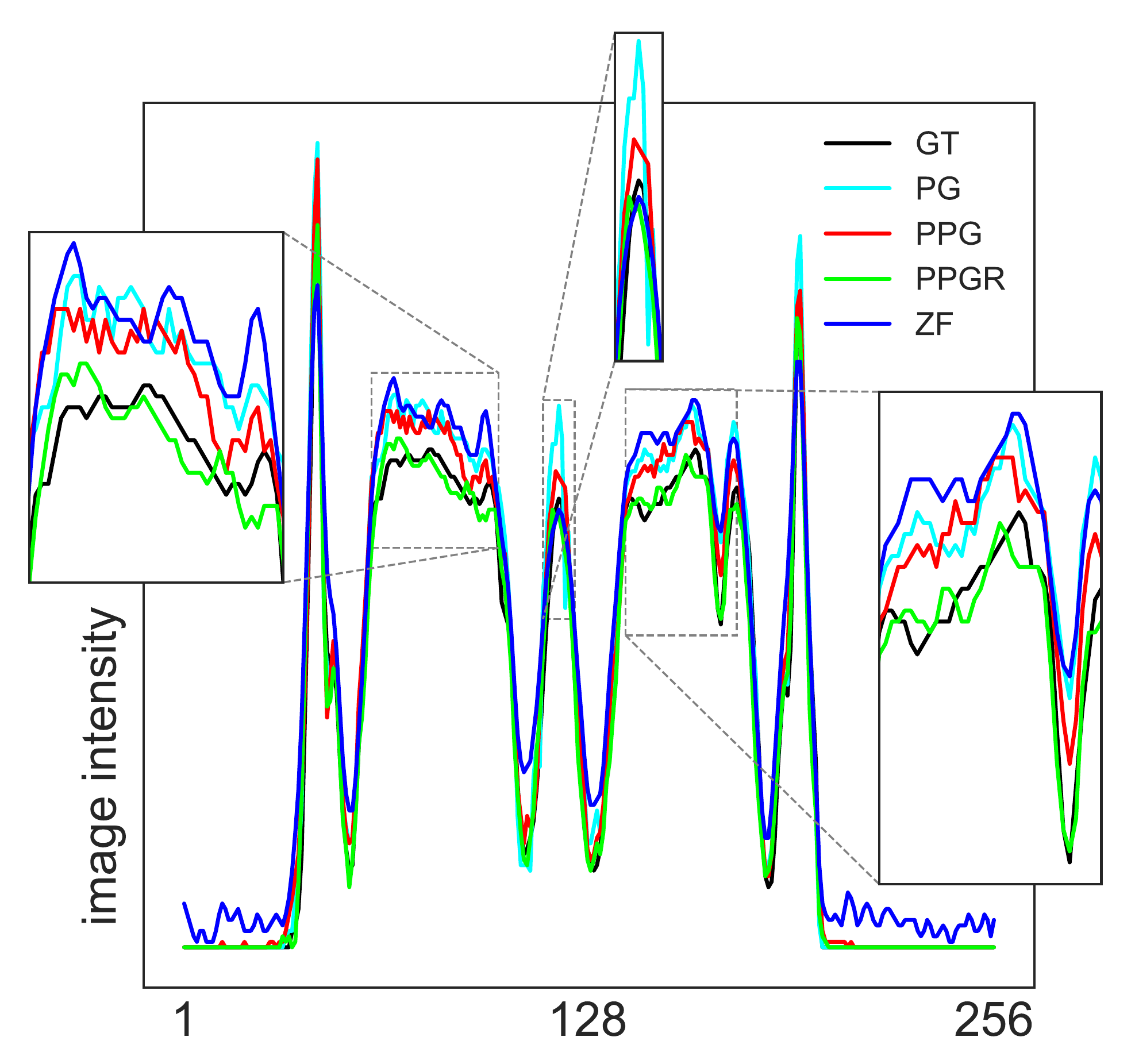}
        \caption{}
    \end{subfigure}%
    ~ 
    \begin{subfigure}[b]{0.5\textwidth}
        \includegraphics[height=0.92in]{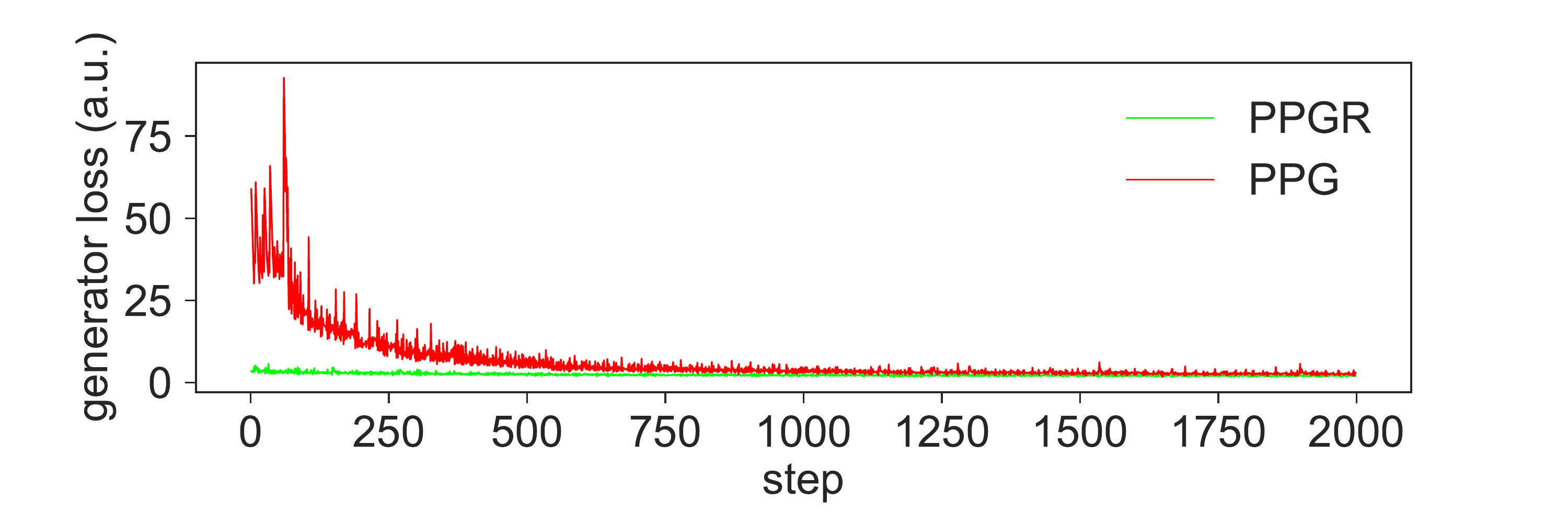}
        \caption{}
        \vspace{1ex}
        \includegraphics[height=0.92in]{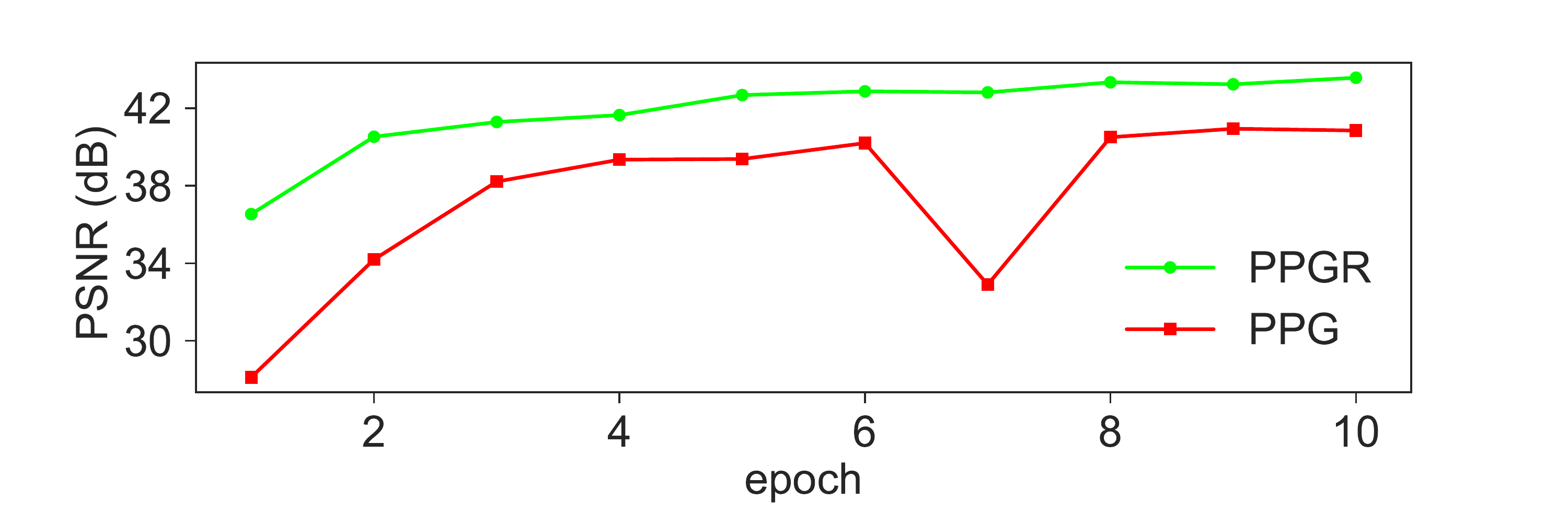}
        \caption{}
    \end{subfigure}
    \caption{\small\bf\it (a) Line profile comparison (red horizontal line in the GT image as shown in Figure \ref{fig:Quali_CS-MRI}), (b) Convergence speed comparison w.r.t. the generator loss and (c) Stability comparison w.r.t. the PSNR of the MICCAI dataset.}
\label{fig:Quant_Convergence}
\end{figure}

\begin{figure}[h]
	\begin{center}
	\includegraphics[scale=0.25, trim={0cm 0cm 0cm 0cm},clip]{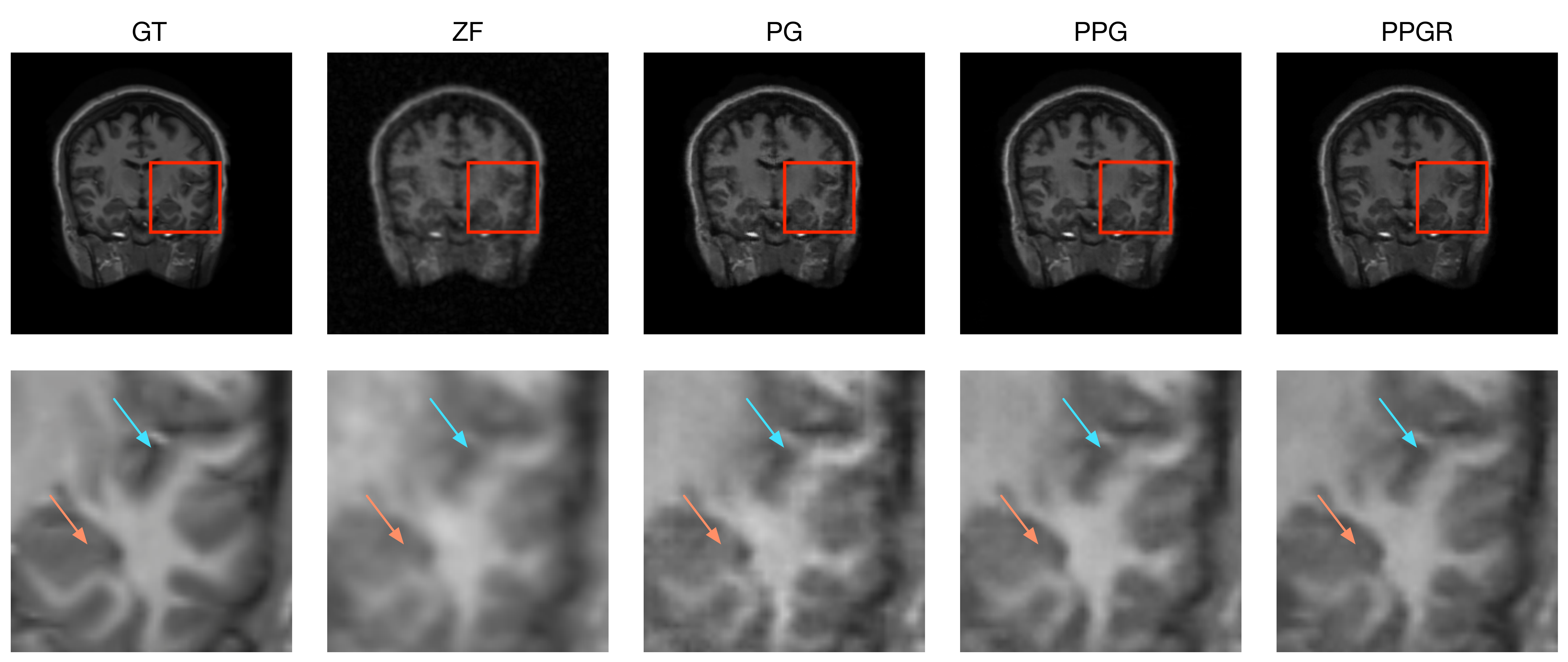}
	\end{center}
	\caption{\small\bf\it Qualitative comparison for our proposed models using 20\% 2D Gaussian random undersampling.}
\label{fig:Quali_CS-MRI_2}
\end{figure}

Compared to previous studies based on deep learning (CNNs) for fast CS-MRI, our conditional GAN-based method also incorporated content loss and refinement learning. The refinement learning can stabilise and speed up our training procedure (Figure \ref{fig:Quant_Convergence} (b) and (c)), while the perceptual loss can make our reconstruction more realistic. Interestingly, we can observe that our PG method without perceptual loss may yield smaller NMSE and higher PSNR than the PPG and PPGR methods that utilised the perceptual loss when the undersampling rate is $\leq$30\% (Table \ref{tab:MICCAI_quant}). This means quantitatively the PG method has outperformed the PPG and PPGR methods; however, the qualitative visualisation demonstrated that results from both PPG and PPGR are superior to PG results in terms of finer perceptual details and less jagged artefacts (Figure \ref{fig:Quali_CS-MRI_2}). This can be attributed to the fact that when reconstructing a highly undersampled \textit{k}-space, the PG method without perceptual loss can only find an optimal solution to satisfy the MSE criteria but may not perceptually resemble the real data. More importantly, conventional or recently proposed deep learning-based fast CS-MRI methods solve the inverse problem by assuming the correctness of the `forward model'. In contrast, our GAN-based method could still perform well without such assumptions imparted by the forward model.

Our method has a similar architecture as \cite{Ledig2016}, which also contains a perceptual loss for solving SR. Essentially, CS-MRI is a more general inverse problem to recover data from undersampled measurements, in which the undersampling pattern is random and noise and artefacts propagation is global due to frequency domain operation (compared to the regular downsampling pattern and local artefacts in SR). Therefore, the CS-MRI is a more challenging problem to solve. On the other hand, our model can be generalised to solve SR. Compared to \cite{Ledig2016}, our PPGR method applied U-Net architecture to reconstruct better image details. In addition, our model is reliable when the raw data is randomly and highly undersampled using the proposed refinement learning. 
\begin{table}[h]
\centering
\caption{\footnotesize\emph{NMSE and PSNR of the MICCAI dataset using different random undersampling masks and ratios.}}
\scalebox{0.88}{
\begin{tabular}{lccccccccccc}
\toprule
\multicolumn{1}{c}{\multirow{2}{*}{Mask}} & \multicolumn{1}{l}{\multirow{2}{*}{Method}} & \multicolumn{2}{c}{10\%} & \multicolumn{2}{c}{20\%} & \multicolumn{2}{c}{30\%} & \multicolumn{2}{c}{40\%} & \multicolumn{2}{c}{50\%} \\ \cmidrule{3-12}
\multicolumn{1}{c}{}                      & \multicolumn{1}{l}{}                        & NMSE        & PSNR       & NMSE        & PSNR       & NMSE        & PSNR       & NMSE        & PSNR       & NMSE        & PSNR       \\ \midrule
\multirow{4}{*}{Gaussian 1D}              & ZF                                          & 0.3297      & 27.26      & 0.1776      & 33.86      & 0.1513      & 35.32      & 0.1186      & 37.49      & 0.0828      & 40.65      \\
                                          & PG                                          & 0.1911      & 33.16      & \textbf{0.0949}      & \textbf{39.29}      & \textbf{0.0875}      & \textbf{40.01}      & 0.0599      & 43.33      & 0.0530      & 44.41      \\
                                          & PPG                                         & 0.2321      & 31.51      & 0.1004      & 38.91      & \textbf{0.0875}      & 39.99      & 0.0695      & 42.31      & 0.0461      & 45.66      \\
                                          & PPGR                                        & \textbf{0.1901}      & \textbf{33.24}      & 0.0958      & 39.22      & 0.0906      & 39.73      & \textbf{0.0528}      & \textbf{44.49}      & \textbf{0.0385}      & \textbf{47.30}      \\ \midrule
\multirow{4}{*}{Gaussian 2D}              & ZF                                          & 0.2374      & 31.25      & 0.1797      & 33.78      & 0.1315      & 36.57      & 0.0981      & 39.16      & 0.0708      & 41.99      \\
                                          & PG                                          & \textbf{0.0890}      & \textbf{39.91}      & \textbf{0.0781}      & \textbf{41.03}      & \textbf{0.0569}      & \textbf{43.89}      & 0.0519      & 44.68      & 0.0371      & 47.64      \\
                                          & PPG                                         & 0.0925      & 39.53      & 0.0878      & 40.12      & 0.0612      & 43.16      & 0.0555      & 44.12      & 0.0403      & 46.85      \\
                                          & PPGR                                        & 0.1029      & 38.61      & 0.0804      & 40.82      & 0.0586      & 43.56      & \textbf{0.0440}      & \textbf{46.20}      & \textbf{0.0347}      & \textbf{48.32}      \\  \bottomrule
\end{tabular}
	\label{tab:MICCAI_quant}%
}
\end{table}
We notice that our current study may have two limitations, but will not influence the final conclusion: (1) we compared to the results in \cite{Yang2016} without implementation. This is because the implementation of those methods is difficult as the fine-tuned hyper-parameters used are not always clearly stated and the methodologies cannot be reproduced exactly. However, we validated our methods using the test cases randomly selected from the same datasets as \cite{Yang2016} that can achieve a relatively fair comparison; (2) the frequency domain information, which may provide useful constraints for reconstruction fidelity, is not explicitly considered in our current model. This is now considered as a future working direction.

\section{Conclusion}
\label{Conclusion}

In this study, we proposed a conditional GAN-based deep learning method to solve the de-aliasing for fast CS-MRI. Remarkably, by incorporating the adversarial loss with a content loss that consists of pixel-wise MSE and perceptual VGG loss, our method can achieve promising and realistic MRI reconstruction results. By using the refinement learning, our method is fast and robust even when the \textit{k}-space raw data is highly undersampled. Convincing simulation based results show promise of our technique to be translational for real MRI acquisition applications.

\bibliographystyle{ieeetr}
\bibliography{CS_MRI_References}

\end{document}